\pdfoutput=1

\documentclass[11pt]{article}

\usepackage[]{emnlp2021}
\usepackage{enumitem}
\usepackage{times}
\usepackage{latexsym}
\usepackage{xspace}
\usepackage[T1]{fontenc}
\usepackage{graphicx}
\usepackage[utf8]{inputenc}

\usepackage{microtype}
\usepackage[utf8]{inputenc} 
\usepackage[T1]{fontenc}    
\usepackage{hyperref}       
\usepackage{url}            
\usepackage{booktabs,colortbl}       
\usepackage{amsfonts}       
\usepackage{nicefrac}       
\usepackage{microtype}      
\usepackage{xcolor}         
\usepackage{CJKutf8}
\usepackage{balance}
\usepackage{bm}
\usepackage{amssymb}
\usepackage{url}
\usepackage{makecell}
\usepackage{colortbl}
\usepackage{textcomp}
\usepackage{CJK}
\usepackage{enumitem}
\usepackage{amsfonts}
\usepackage{caption}
\usepackage{float}
\usepackage{multirow}
\usepackage{color}
\usepackage{arydshln}
\usepackage{wrapfig}
\usepackage{graphicx}
\usepackage{amsmath}
\usepackage{enumitem}
\usepackage{graphics}
\usepackage{graphicx}
\usepackage{subfig}

\usepackage{url}            
\usepackage{booktabs}       
\usepackage{amsfonts}       
\usepackage{nicefrac}       
\usepackage{microtype}      
\usepackage{amsmath}
\usepackage{makecell}
\usepackage{caption}
\usepackage{float}
\usepackage{multirow}
\usepackage{color}
\usepackage{arydshln}
\definecolor{liam}{RGB}{220,220,220}
\DeclareMathOperator*{\argmax}{arg\,max}
\renewcommand\vec[1]{\overrightarrow{#1}}
\newcommand\cev[1]{\overleftarrow{#1}}
\newcommand\bid[1]{\overleftrightarrow{#1}}

\title{Improving Neural Machine Translation by Bidirectional Training}

\author{
Liang Ding\\
The University of Sydney\\
\normalsize \texttt{ldin3097@sydney.edu.au}\\
\And
Di Wu
\\
Peking University\\
\normalsize \texttt{inbath@163.com}\\
\And
Dacheng Tao\\
JD Explore Academy, JD.com\\
\normalsize \texttt{dacheng.tao@gmail.com}}

\begin{document}
\maketitle
\begin{abstract}
We present a simple and effective pretraining strategy -- bidirectional training (BiT) for neural machine translation. Specifically, we bidirectionally update the model parameters at the early stage and then tune the model normally. To achieve bidirectional updating, we simply reconstruct the training samples from ``src$\rightarrow$tgt'' to  ``src+tgt$\rightarrow$tgt+src'' without any complicated model modifications.  Notably, our approach does not increase any parameters or training steps, requiring the parallel data merely. Experimental results show that BiT pushes the SOTA neural machine translation performance across 15 translation tasks on 8 language pairs (data sizes range from 160K to 38M) significantly higher. Encouragingly, our proposed model can complement existing data manipulation strategies, i.e. back translation, data distillation and data diversification. Extensive analyses show that our approach functions as a novel bilingual code-switcher, obtaining better bilingual alignment.
\end{abstract}

\section{Introduction}
Recent years have seen a surge of interest in neural machine translation (NMT,~\citealp{luong2015effective,wu2016google,gehring2017convolutional,transformer}) where it benefits from a massive amount of training data. But obtaining such large amounts of parallel data is not-trivial in most machine translation scenarios. For example, there are many low-resource language pairs (e.g. English-to-Tamil), which lack adequate parallel data for training. 

Although many approaches about fully exploiting the parallel and monolingual data are proposed, e.g. back translation~\cite{sennrich-etal-2016-improving}, knowledge distillation~\cite{kim-rush-2016-sequence} and data diversification~\cite{nguyen2019data}, the prerequisite of these approaches is to build a well-performed baseline model based on the parallel data. However, \newcite{koehn2017six,lample2018phrase,sennrich2019revisiting} empirically reveal that NMT runs worse than their statistical or even unsupervised counterparts in low-resource conditions. Here naturally arise a question: \textit{Can we find a strategy to consistently improve NMT performance given the parallel data merely?}

We decide to find a solution from \textit{human learning behavior}.~\newcite{pavlenko2002bidirectional,dworin2003insights,chen2015bilingual} show that bidirectional language learning helps master bilingualism. In the context of machine translation, both the source$\rightarrow$target and target$\rightarrow$source language mappings may benefit bilingual modeling, which motivates many recent studies, e.g. dual learning~\cite{he2016dual} and symmetric training~\cite{cohn2016incorporating,Liang2007AgreementBasedL}. However, their approaches rely on external resources (e.g. word alignment or monolingual data) or complicated model modifications, which limit the applicability of the method to a broader range of languages and model structures. Accordingly, we turn to propose a simple data manipulation strategy and transfer the bidirectional relationship through \textit{bidirectional training} (\S\ref{subsec:bidirectional}). The core idea is using a bidirectional system as an initialization for a unidirectional system.
Specifically, to make the most of the parallel data, we first reconstruct the training samples from ``$\vec{\text{B}}$: source$\rightarrow$target'' to ``$\bid{\text{B}}$: source$+$target$\rightarrow$target$+$source'', where the training data was doubled. Then we update the model parameters with $\bid{\text{B}}$ in the early stage, and tune the model with normal ``$\vec{\text{B}}$ source$\rightarrow$target'' direction.

We validated our approach on several benchmarks across different language families and data sizes, including IWSLT21 En$\leftrightarrow$De, WMT16 En$\leftrightarrow$Ro, 
WMT19 En$\leftrightarrow$Gu,
IWSLT21 En$\leftrightarrow$Sw, WMT14 En$\leftrightarrow$De, WMT19 En$\leftrightarrow$De, WMT17 Zh$\leftrightarrow$En and WAT17 Ja$\leftrightarrow$En. Experimental results show that the proposed bidirectional training (BiT) consistently and significantly improves the translation performance over the strong Transformer~\cite{transformer}. 
Also, we show that BiT can complement existing data manipulation strategies, i.e. back translation, knowledge distillation and data diversification. Extensive analyses in \S\ref{subsec:analysis} confirm that the performance improvement indeed comes from the better cross-lingual modeling and our method works like a novel code-switching method.

\section{Bidirectional Training}
\subsection{Preliminary}
\label{subsec:standard}
Given a source sentence $\bf x$, an NMT model generates each target word ${\bf y}_t$ conditioned on previously generated ones ${\bf y}_{<t}$. Accordingly, the probability of generating $\bf y$ is computed as:
\begin{equation}
    p({\bf y}|{\bf x})
    =\prod_{t=1}^{T}p({\bf y}_t|{\bf x},{\bf y}_{<t}; \theta)
    \label{eq:standard}
\end{equation}
where $T$ is the length of the target sequence and the parameters $\theta$ are trained to maximize the likelihood of a set of training examples according to $\mathcal{L}(\theta) = \argmax_{\theta} \log p({\bf y}|{\bf x}; \theta)$. Typically, we choose Transformer~\cite{transformer} as its SOTA performance. The training examples can be formally defined as follows:
\begin{equation}
    \vec{\text{B}} = \{(\mathbf{x}_i, \mathbf{y}_i)\}^N_{i=1}
    \label{eq:B}
\end{equation}
where $N$ is the total number of sentence pairs in the training data. Note that in standard MT training, the $\bf{x}$ is fed into the encoder and $\textbf{y}_{<t}$ into the decoder to finish the conditional estimation for $\textbf{y}_t$, thus the utilization of $\vec{\text{B}}$ is directional, i.e. $\mathbf{x}_i$$\rightarrow$$\mathbf{y}_i$.

\subsection{Pretraining with Bidirectional Data}
\label{subsec:bidirectional}
\paragraph{Motivation}
The motivation is when human learn foreign languages with translation examples, e.g. $\mathbf{x}_i$ and $\mathbf{y}_i$. Both directions of this example, i.e. $\mathbf{x}_i$$\rightarrow$$\mathbf{y}_i$ and $\mathbf{y}_i$$\rightarrow$$\mathbf{x}_i$, may help human easily master the bilingual knowledge. 
Motivated by this, \newcite{levinboim-etal-2015-model,Liang2007AgreementBasedL} propose to modelling the invertibility between bilingual languages. \newcite{cohn2016incorporating} introduce extra bidirectional prior regularization to achieve symmetric training from the point view of training objective. \newcite{he2018layer,zheng2019mirror,ding2020acl} enhance the coordination of bidirectional corpus with model level modifications. Different from above methods, we model both directions of a given training example by a simple data manipulation strategy.

\paragraph{Our Approach}
Many studies have shown that pretraining could transfer the knowledge and data distribution, hence improving the generalization~\cite{hendrycks2019using,mathis2021pretraining}. Here we want to transfer the bidirectional knowledge among the corpus. Specifically, we propose to first pretrain MT models on bidirectional corpus, which can be defined as follows:
\begin{equation}
    \bid{\text{B}} = \{(\mathbf{x}_i, \mathbf{y}_i) \cup (\mathbf{y}_i, \mathbf{x}_i)\}^N_{i=1}
    \label{eq:bi.B}
\end{equation}
such that the $\theta$ in Equation~\ref{eq:standard} can be updated by both directions. Then the bidirectional pretraining objective can be formulated as:
\begin{align}
    \bid{\mathcal{L}}(\theta) = &\overbrace{\argmax_{\theta} \log p({\bf y}|{\bf x}; \theta)}^{\text{Forward}: \vec{\mathcal{L}_{\theta}}}\\
    &+ \underbrace{\argmax_{\theta} \log p({\bf x}|{\bf y}; \theta)}_{\text{Backward}: \cev{\mathcal{L}_{\theta}}}
    \label{eq:bi.B}
\end{align}
where the $\text{forward}~\vec{\mathcal{L}_{\theta}}$ and $\text{backward}~\cev{\mathcal{L}_{\theta}}$ are optimized iteratively. 

From data perspective, we achieve the bidirectional updating as follows: 1) swapping the source and target sentences of a parallel corpus, and 2) appending the swapped data to the original. Then the training data was doubled to make better and full use of the costly bilingual corpus.
The pretraining can acquire general knowledge from bidirectional data, which may help {\em better} and {\em faster} learning further tasks. 
Thus, we early stop BiT at 1/3 of the total training steps (we discuss its reasonability in \S\ref{subsec:setup}). In order to ensure the proper training direction, we further train the pretrained model on required direction $\vec{\text{B}}$ with the rest of 2/3 training steps.
Considering the effectiveness of pretraining~\cite{mathis2021pretraining} and clean finetuning~\cite{wu2019exploiting}, we introduce a combined pipeline: $\bid{\text{B}}\rightarrow\vec{\text{B}}$ as out best training strategy. There are many possible ways to implement the general idea of bidirectional pretraining. The aim of this paper is not to explore the whole space but simply to show that one fairly straightforward implementation works well and the idea is reasonable.

\begin{table*}[t]
    \centering
    \setlength{\tabcolsep}{3.2pt}
    \scalebox{0.98}{
    \begin{tabular}{lllllllllllc}
    \toprule
    \textbf{Data Source} &
    \multicolumn{2}{c}{\textbf{IWSLT14}} &
    \multicolumn{2}{c}{\textbf{WMT16}}  &
    \multicolumn{2}{c}{\textbf{IWSLT21}} & 
    \multicolumn{2}{c}{\textbf{WMT14}} & 
    \multicolumn{2}{c}{\textbf{WMT19}} &
    \multirow{2}{*}{$\Delta$}\\
    \textbf{Size} &
    \multicolumn{2}{c}{160K} &
    \multicolumn{2}{c}{0.6M} &
    \multicolumn{2}{c}{2.4M} &
    \multicolumn{2}{c}{4.5M} &
    \multicolumn{2}{c}{38M} &\\
    \cdashline{2-11}
    \textbf{Direction} & \bf En-De &\bf De-En & \bf En-Ro & \bf Ro-En  & \bf En-Sw & \bf Sw-En & \bf En-De &\bf De-En &\bf En-De &\bf De-En& \textbf{\textit{Ave.}}\\
    \midrule
    {\bf Transformer} & 29.2 & 35.1  & 33.9& 34.1  & 28.8 & 48.5 & 28.6 & 32.1 & 39.9 & 40.1 &--\\
    {\bf ~~~~+BiT} & 29.9$^\dagger$ & 36.3$^\ddagger$ & 35.2$^\ddagger$ & 35.9$^\ddagger$ & 29.9$^\ddagger$ & 49.9$^\ddagger$ & 29.7$^\ddagger$ & 32.9$^\dagger$ & 40.5$^\dagger$ & 41.6$^\ddagger$ & +1.1\\
    \bottomrule
    \end{tabular}}
    \caption{Comparison with previous AT work on several widely-used benchmarks, including IWSLT14 En$\leftrightarrow$De, WMT16 En$\leftrightarrow$Ro, IWSLT21 En$\leftrightarrow$Sw, WMT14 En$\leftrightarrow$De and WMT19 En$\leftrightarrow$De. ``$^{\ddagger/\dagger}$'' indicates significant difference ($p < 0.01/ 0.05$) from corresponding baselines, and this leaves as default symbol in Table~\ref{tab:distantlanguage}-\ref{tab:extreme-low}.}
    \label{tab:main-results}
\end{table*}
\section{Experiments}
\subsection{Setup}
\label{subsec:setup}
\paragraph{Data}
Main experiments in Table~\ref{tab:main-results} are conducted on five translation datasets: IWSLT21 English$\leftrightarrow$German~\cite{nguyen2019data}, WMT16 English$\leftrightarrow$Romania~\cite{gu2018non}, IWSLT21 English$\leftrightarrow$Swahili\footnote{\url{https://iwslt.org/2021/low-resource}}, WMT14 English$\leftrightarrow$German~\cite{transformer} and WMT19 English$\leftrightarrow$German\footnote{\url{http://www.statmt.org/wmt19/translation-task.html}}. The data sizes can be found in Table~\ref{tab:main-results}, ranging from 160K to 38M. 
Two distant language pairs in Table~\ref{tab:distantlanguage} are WMT17 Chinese$\leftrightarrow$English~\cite{hassan2018achieving} and WAT17 Japanese$\rightarrow$English~\cite{morishita2017ntt}, containing 20M and 2M training examples, respectively. The monolingual data used for back translation in Table~\ref{tab:complementary} is randomly sampled from publicly available News Crawl corpus\footnote{\url{http://data.statmt.org/news-crawl/}}.
We use same valid\& test sets with previous works for fair comparison except IWSLT21 English$\leftrightarrow$Swahili, where we follow \citet{Ding2021iwslt} to sample 5K/ 5K sentences from the training set as valid/ test sets.
We preprocess all data via BPE~\cite{Sennrich:BPE} with 32K merge operations. We use tokenized BLEU~\cite{papineni2002bleu} as the evaluation metric for all languages except English$\rightarrow$Chinese, where we use SacreBLEU\footnote{BLEU+case.mixed+lang.en-zh+numrefs.1+smooth.exp+test.wmt17+tok.zh+version.1.5.1}~\cite{post-2018-call}. The \textit{sign-test}~\cite{collins2005clause} is used for statistical significance test.

\paragraph{Model}
We validated our proposed BiT on Transformer~\cite{transformer}\footnote{\url{https://github.com/pytorch/fairseq}}. All language pairs are trained on Transformer-\textsc{Big} except IWSLT14 En$\leftrightarrow$De and WMT16 En$\leftrightarrow$Ro (trained on Transformer-\textsc{Base}) because of their extremely small data size. For fair comparison, we set beam size and length penalty as 5 and 1.0 for all language pairs.
It is worth noting that our data-level approach neither modifies model structure nor adds extra training loss, thus it's feasible to deploy on any frameworks, e.g. DynamicConv~\cite{wu2019pay} and non-autoregressive MT~\cite{gu2018non,Ding2020coling,ding2021iclr}, and training orders, e.g. curriculum learning~\cite{liu2020acl,zhou2021self,zhan2021meta,Ding2021acl2}. We will explore them in the future works. 

\paragraph{Training}
For Transformer-\textsc{Big} models, we empirically adopt large batch strategy~\cite{edunov2018understanding} (i.e. 458K tokens/batch) to optimize the performance. The learning rate warms up to $1\times10^{-7}$ for 10K steps, and then decays for 30K (data volumes range from 2M to 10M) / 50K (data volumes large than 10M) steps with the cosine schedule; For Transformer-\textsc{Base} models, we empirically adopt 65K tokens per batch for small data sizes, e.g. IWSLT14 En$\rightarrow$De and WMT16 En$\rightarrow$Ro. The learning rate warms up to $1\times10^{-7}$ for 4K steps, and then decays for 26K steps.
For regularization, we tune the dropout rate from [0.1, 0.2, 0.3] based on validation performance, and apply weight decay with
0.01 and label smoothing with $\epsilon$ = 0.1. We use Adam optimizer ~\citep{kingma2015adam} to train the models. We evaluate the performance on the averaged last 10 checkpoints to avoid stochasticity.

Someone may doubt that BiT heavily depends on how to properly set the early-stop steps. To dispel the doubt, we investigate whether our approach is robust to different early-stop steps. In preliminary experiments, we tried several simple fixed early-stop steps according to the size of training data (e.g. training 40K En-De and early stop at 10K/ 15K/ 20K, respectively). We found that both strategies achieve similar performances. Thus, we decide to choose a simple and effective method (i.e. 1/3 of the total training steps) for better reprehensibility.

\subsection{Results}
\paragraph{Results on Different Data Scales}
To confirm the effectiveness of our method across different data sizes,
we experimented on 10 language directions, including IWSLT14 En$\leftrightarrow$De, WMT16 En$\leftrightarrow$Ro, IWSLT21 En$\leftrightarrow$Sw, WMT14 En$\leftrightarrow$De and WMT19 En$\leftrightarrow$De. The smallest one merely contains 160K sentences, while the largest direction includes 38M sentence pairs. Table~\ref{tab:main-results} reports the results, we show that BiT achieves significant improvements over strong baseline Transformer in 7 out of 10 directions under the significance test $p<0.01$, and the rest of 3 directions also show promising performance under the significance test $p<0.05$, demonstrating the effectiveness and universality of our proposed bidirectional pretraining strategy.  
Notably, one advantage of BiT is it saves 1/3 of the training time for the reverse direction. For example, the pretrained BiT checkpoint for En$\rightarrow$De can be used to tune the reverse direction De$\rightarrow$En. This advantage shows BiT could be an efficient training strategy for multilinguality, e.g. multi-lingual pretraining~\cite{Liu2020mbart} and translation~\cite{hatoward}.

\paragraph{Results on Distant Language Pairs}
Inspired by \citet{Ding2021acl1}, to dispel the doubt that BiT could merely be applied on languages within the same language family, e.g. English and German, we report the results of BiT on Zh$\leftrightarrow$En and Ja$\rightarrow$En language pairs, which belong to different language families (i.e. Indo-European, Sino-Tibetan and Japonic). 

Table~\ref{tab:distantlanguage} lists the results, as seen, compared with baselines, our method significantly and incrementally improves the translation quality in all cases. In particular, BiT achieves averaged +0.9 BLEU improvement over the baselines, showing the effectiveness and universality of our method across language pairs.

\begin{table}[t]
    \centering
    \scalebox{1}{
    \begin{tabular}{lccc}
    \toprule
    \bf Data Source &   \multicolumn{2}{c}{\bf WMT17}   &   \multicolumn{1}{c}{\bf WAT17}  \\
    \bf Size    & \multicolumn{2}{c}{20M} &  \multicolumn{1}{c}{2M}\\
    \cdashline{2-3} \cdashline{4-4}
    \bf Direction  & \bf Zh-En &\bf En-Zh &\bf Ja-En\\
    \midrule
    {\bf Transformer} & 23.7 & 33.2 & 28.1 \\
    \bf {\bf ~~~~+BiT}& 24.9$^\ddagger$ & 33.9$^\dagger$ & 28.8$^\dagger$ \\
    \bottomrule 
    \end{tabular}}
    \caption{Performance on distant language pairs, including WMT17 Zh$\leftrightarrow$En and WAT17 Ja$\rightarrow$En. To perform BiT on languages in different alphabets, we share the sub-words dictionaries between languages.}
    \label{tab:distantlanguage}
\end{table}

\begin{table}[t]
    \centering
    \scalebox{0.95}{
    \begin{tabular}{lc}
    \toprule
    \bf Model & \bf BLEU \\
    \midrule
    \textbf{Transformer-\textsc{Big}}/+BiT & 28.6/ 29.7$^\ddagger$\\
    \hdashline
    \textbf{~~+BT}\cite{caswell-etal-2019-tagged}/+BiT & 30.5/ 31.2$^\dagger$\\
    \hdashline
    \textbf{~~+KD}\cite{kim-rush-2016-sequence}/+BiT & 29.3/ 30.1$^\dagger$\\
    \hdashline
    \textbf{~~+DD}\cite{nguyen2019data}/+BiT & 30.1/ 30.7$^\dagger$\\
    \bottomrule
    \end{tabular}}
\caption{Complementary to other works. ``/+BiT'' means combining BiT with corresponding works, and BLEU scores of BiT followed their counterparts with ``/''. Experiments are conducted on WMT14 En-De.}
\label{tab:complementary}
\end{table}

\paragraph{Complementary to Related Work}
Recent studies start to combine pretraining and traditional data manipulation approaches for better model performance~\citep{lample2019cross,Liu2020mbart,liu2021emnlp}. To show the complementary between our proposed pretraining method BiT and related data manipulation works,
we list three representative data manipulation approaches for NMT: a) {Tagged Back Translation} (\textbf{BT},~\citealt{caswell-etal-2019-tagged}) combines the synthetic data generated with target-side \textit{monolingual data} and parallel data; b) {Knowledge Distillation} (\textbf{KD},~\citealt{kim-rush-2016-sequence}) trains the model with sequence-level distilled \textit{parallel data}; c) {data diversification} (\textbf{DD},~\citealt{nguyen2019data}) 
diversifies the data by applying KD and BT on \textit{parallel data}. 
As seen in Table~\ref{tab:complementary}, BiT can be applied on existing data manipulation approaches and yield further significant improvements. 

\subsection{Analysis}
\label{subsec:analysis}
We conducted analyses to better understand BiT. Unless otherwise stated, all results are reported on the WMT14 En-De.
\begin{table}[tb]
    \centering
    \scalebox{1}{
    \begin{tabular}{ll}
        \toprule
         \textbf{Model} & \bf BLEU  \\
        \midrule
         \bf Transformer-\textsc{Big}&28.6\\
         {\bf ~~~+mRASP}~\cite{lin2020pre}&29.3$^\dagger$\\
         {\bf ~~~+CSP}~\cite{yang2020csp}&29.4$^\dagger$\\
         {\bf ~~~+BiT} (Ours)&29.7$^\ddagger$\\
        \bottomrule
    \end{tabular}}
    \caption{Comparison with previous code-switch approaches on bilingual data, where we follow the best settings of ``+mRASP'' and ``+CSP'' as default without extra parameter tuning. For fair comparison, the pretrain/ finetune steps are identical with ours.}
    \label{tab:codeswitch}
\end{table}
\paragraph{BiT works as a simple bilingual code-switcher}
\newcite{lin2020pre,yang2020csp} employ the third-party tool to obtain the alignment information to perform code-switching pertraining, where partial of the source tokens is replaced with the aligned target ones. But training such alignment model is time-consuming and the alignment errors may be propagated. 
Actually, BiT can be viewed as a novel yet simple bilingual code-switcher, where the switch span is the whole sentence and both the source- and target-side sentences are replaced with the probability $0.5$. 
\begin{CJK}{UTF8}{gbsn}
Take a sentence pair \{``Bush held a talk with Sharon''$\rightarrow$``布什 与 沙龙 举行 了 会谈''\} in English$\rightarrow$Chinese dataset as an example, during pretraining phase, the reconstructed corpus contains \{``Bush held a talk with Sharon'' $\rightarrow$ ``布什 与 沙龙 举行 了 会谈''\} and its reversed version {``布什 与 沙龙 举行 了 会谈'' $\rightarrow$ ``Bush held a talk with Sharon''}, simultaneously. For the English$\rightarrow$Chinese direction, the reversed sentence pair exactly belongs to the sentence-level switch with a probability of 0.5. \end{CJK}
For fair comparison, we implement \citealp{lin2020pre,yang2020csp}'s approaches in \textit{bilingual data scenario}. Table~\ref{tab:codeswitch} show the superiority of BiT, indicating BiT is a good alternative to code-switch in bilingual scenario.

\paragraph{BiT improves alignment quality}
Our proposed BiT intuitively encourages self-attention to learn bilingual agreement, thus has the potential to induce better attention matrices. 
We explore this hypothesis on the widely-used Gold Alignment dataset\footnote{\url{http://www-i6.informatik.rwth-aachen.de/goldAlignment}, the original dataset is German-English, we reverse it to English-German.} and follow~\citet{tang2019understanding} to perform the alignment. The only difference being that we average the attention matrices across all heads from the penultimate layer~\cite{garg2019jointly}. The alignment error rate (AER,~\citealt{och2003systematic}), precision (P) and recall (R) are evaluation metrics. Table~\ref{tab:align} summarizes that BiT allows Transformer to learn better attention matrices, thereby improving alignment performance (24.3 vs. 27.1).
\begin{table}[tb]
    \centering
    \scalebox{0.97}{
    \begin{tabular}{lccc}
        \toprule
         \textbf{Model}&{\textbf{AER}}&\textbf{P}&\textbf{R}  \\
        \midrule
         \bf Transformer-\textsc{Big}&27.1\%&71.2\%&74.7\%\\
         {\bf ~~~~+BiT}&24.3\%&74.6\%&76.9\%\\
        \bottomrule
    \end{tabular}}
    \caption{The AER scores of alignments on En-De.}
    \label{tab:align}
\end{table}

\begin{table}[tb]
    \centering
    \scalebox{1}{
    \begin{tabular}{lccc}
        \toprule
         \textbf{Model}&{\textbf{En$\rightarrow$Gu}}&\textbf{Gu$\rightarrow$En}&\textbf{Ave.$\Delta$}  \\
        \midrule
         \bf Base & 3.0 & 8.2 & -\\
         \rowcolor{liam}{\bf ~~~~+BT}& 2.6 & 10.1 & - \\
         \hdashline
         \bf Base+BiT & 4.2$^\ddagger$ & 9.0$^\ddagger$ & +1.0\\
         \rowcolor{liam}{\bf ~~~~+BT}& 5.8$^\ddagger$ & 12.4$^\ddagger$ & +2.8\\
        \bottomrule
    \end{tabular}}
    \caption{Results for En$\leftrightarrow$Gu on WMT2019 test sets. ``Ave. $\Delta$'' shows the averaged improvements of ``Base+BiT'' v.s. ``Base'' and their corresponding ``+BT'' comparisons.}
    \label{tab:extreme-low}
\end{table}

\paragraph{BiT works for extremely low-resource settings} 
Researches may doubt BiT may fail on extremely low-resource settings where \textit{back-translation even does not work}. To dispel this concern, we conduct experiments on WMT19 English$\leftrightarrow$Gujurati\footnote{\url{http://www.statmt.org/wmt19/translation-task.html}} in Table~\ref{tab:extreme-low}. Specifically, we follow \citet{li2019niutrans} to collect and preprocess the parallel data to build the base model ``Base'' and our ``Base+BiT'' model. For a fair comparison, we sample the monolingual English and Gujurati sentences to ensure Parallel: Monolingual = 1:1 to generate the synthetic data. As seen, when directly applying back-translation (BT) on the En$\leftrightarrow$Gu Base model, there indeed shows a slight performance drop (-0.4 BLEU). However, our ``BiT'' significantly improves the initial Base model by averaged +1.0 BLEU, and making the BiT-equipped BT more effective compared to vanilla BT (+2.8 BLEU). 
These findings on extremely low-resource settings demonstrate that 1) our BiT consistently works well; and 2) BiT provides a better initial model, thus rejuvenating the effects of back-translation.

\section{Conclusion and Future Works}
In this study, we propose a pretraining strategy for NMT with parallel data merely. Experiments show that our approach significantly improves translation performance, and can complement existing data manipulation strategies. Extensive analyses reveal that our method can be viewed as a simple yet better bilingual code-switching approach, and improves bilingual alignment quality. 

Encouragingly, with BiT, our system~\cite{Ding2021iwslt} got the first place in terms of BLEU scores in IWSLT2021\footnote{\url{https://iwslt.org/2021/}} low-resource track. 
It will be interesting to
integrate BiT into our previous systems~\cite{ding2019wmt,Wang2020TencentAL} and validate its effectiveness on industrial level competitions, e.g. WMT\footnote{\url{http://www.statmt.org/wmt21/}}. It is also worthwhile to explore the effectiveness of our proposed bidirectional pretraining strategy on multilingual NMT task~\cite{hatoward}.
\section*{Acknowledgments}
We would thank the anonymous reviewers and the area chair for their considerate proofreading and valuable comments.

\bibliography{custom}
\bibliographystyle{acl_natbib}

\appendix
\end{document}